\documentclass[10pt,twocolumn,letterpaper]{article}

\usepackage{wacv}              

\usepackage{graphicx}
\usepackage{amsmath}
\usepackage{amssymb}
\usepackage{booktabs}

\usepackage{times}
\usepackage{threeparttable}
\usepackage{multirow}
\usepackage{natbib}
\setcitestyle{numbers,square}
\usepackage[utf8]{inputenc} 
\usepackage[T1]{fontenc}    
\usepackage{url}            
\usepackage{amsfonts}       
\usepackage{nicefrac}       
\usepackage{microtype}      
\usepackage{xcolor}         
\usepackage{caption}

\usepackage[accsupp]{axessibility}  

\usepackage[pagebackref,breaklinks,colorlinks]{hyperref}

\usepackage[capitalize]{cleveref}
\crefname{section}{Sec.}{Secs.}
\Crefname{section}{Section}{Sections}
\Crefname{table}{Table}{Tables}
\crefname{table}{Tab.}{Tabs.}

\def\wacvPaperID{1856} 
\def\confName{WACV}
\def\confYear{2024}

\begin{document}

\title{DiffCLIP: Leveraging Stable Diffusion for Language Grounded 3D Classification}


\author{
 \textbf{Sitian Shen}$^{1}$, \textbf{Zilin Zhu}$^1$, \textbf{Linqian Fan}$^2$, \textbf{Harry Zhang}$^3$, \textbf{Xinxiao Wu}$^1$ \\
$^1$Beijing Key Laboratory of Intelligent Information Technology, School of Computer Science \\
and Technology, Beijing Institute of Technology, Beijing, China, \\
$^2$Beijing Jiaotong University, Beijing, China, \\
$^3$Massachusetts Institute of Technology, Cambridge, USA, \\
sitian@bit.edu.cn, smzzlcn@gmail.com, LinqFan@bjtu.edu.cn, harryz@mit.edu, wuxinxiao@bit.edu.cn
}

\maketitle

\begin{abstract}
Large pre-trained models have revolutionized the field of computer vision by facilitating multi-modal learning. Notably, the CLIP model has exhibited remarkable proficiency in tasks such as image classification, object detection, and semantic segmentation. Nevertheless, its efficacy in processing 3D point clouds is restricted by the domain gap between the depth maps derived from 3D projection and the training images of CLIP.

This paper introduces DiffCLIP, a novel pre-training framework that seamlessly integrates stable diffusion with ControlNet. The primary objective of DiffCLIP is to bridge the domain gap inherent in the visual branch. Furthermore, to address few-shot tasks in the textual branch, we incorporate a style-prompt generation module.

Extensive experiments on the ModelNet10, ModelNet40, and ScanObjectNN datasets show that DiffCLIP has strong abilities for 3D understanding. By using stable diffusion and  style-prompt generation, DiffCLIP  achieves  an accuracy of 43.2\% for zero-shot classification on OBJ\_BG of ScanObjectNN, which is state-of-the-art performance, and an accuracy of 82.4\% for zero-shot classification on ModelNet10, which is also state-of-the-art performance. 
\end{abstract}

\section{Introduction}

Significant advancements in Natural Language Processing (NLP), particularly in the context of large pre-trained models, have considerably impacted computer vision, enabling multi-modal learning, transfer learning, and spurring the development of new architectures~\cite{dosovitskiy2020image}. These pre-trained NLP models have been instrumental in the extraction of high-level features from text data, subsequently enabling the fusion with visual features from images or videos, which in turn improves performance on tasks such as image captioning and video classification. Recently, OpenAI's Contrastive Language Image Pre-training (CLIP)~\cite{clip} model has been particularly popular due to its superior performance in image classification, object detection, and semantic segmentation. These achievements underscore the potential of large vision-language pre-trained models across a spectrum of applications.

The remarkable performance of large pre-trained models on 2D vision tasks has stimulated researchers to investigate the potential applications of these models within the domain of 3D point cloud processing. Several approaches, including PointCLIP~\cite{pointclip}, PointCLIP v2~\cite{pointclipv2} and CLIP2Point 
~\cite{huang2022clip2point} have been proposed over the years. While these methods demonstrate advantages in zero-shot and few-shot 3D point cloud processing tasks, experimental results on real-world datasets suggest that their efficacy in handling real-world tasks is limited. This limitation is primarily attributed to the significant domain gap between 2D depth maps derived from 3D projection and training images of CLIP. Consequently, the primary focus of our work is to minimize this domain gap and improve the performance of CLIP on zero-shot and few-shot 3D vision tasks.

Previous studies have made significant progress in addressing this challenge. For instance, ~\cite{CG3D} leverages prompt learning  to adapt the input domain of  the pre-trained visual encoder from computer-generated images of CAD objects to real-world images. The CLIP$^2$~\cite{zeng2023clip} model aligns three-dimensional space with a natural language representation that is applicable in real-world scenarios, facilitating knowledge transfer between domains without prior training. To further boost the performance, we introduce a style feature extraction step and a style transfer module via stable diffusion~\cite{stablediffu} to further mitigate the domain gap.

In this paper, we propose  a novel pre-training framework, called DiffCLIP, to minimize the domain gap within both the visual and textual branches. On the visual side, we design a multi-view projection on the original 3D point cloud, resulting in multi-view depth maps. Each depth map undergoes a style transfer process guided by stable diffusion and ControlNet~\cite{controlnet}, generating photorealistic 2D RGB images. The images generated from the style transfer module are then input into the frozen image encoder of CLIP. For zero-shot tasks, we design handcrafted prompts for both the stable diffusion module and the frozen text encoder of CLIP. For few-shot tasks, we further establish a style-prompt generation module that inputs style features extracted from a pre-trained point-cloud-based network and processed via a meta-net, and outputs the projected features as the style-prompt. Coupled with the class label, the entire prompt is then fed into the frozen text encoder of  CLIP  to guide the downstream process.

To demonstrate the effectiveness of DiffCLIP, we evaluate its zero-shot and few-shot classification ability on both synthetic and real-world datasets like ModelNet and ScanObjectNN.
Our main contributions are summarized as follows: 

\begin{enumerate}
\item We propose DiffCLIP, a novel neural network architecture that combines a pre-trained CLIP model with point-cloud-based networks for enhanced 3D understanding.
\item We develop a technique that effectively minimizes the domain gap in point-cloud processing, leveraging the capabilities of stable diffusion and style-prompt generation, which provides significant improvements in the task of point cloud understanding.
\item We conduct experiments on widely adopted benchmark datasets such as ModelNet and the more challenging ScanObjectNN to illustrate the robust 3D understanding capability of DiffCLIP and conduct ablation studies that evaluate the significance of each component within DiffCLIP's architecture.

\end{enumerate}

\section{Related Work}
\subsection{3D Shape Classification}
3D point cloud processing methods for classification can be divided into three categories: projection-based~\cite{multi-view1, you2018pvnet, lim2021planar}, volumetric-based~\cite{maturana2015voxnet, riegler2017octnet, sim2019personalization} and point-based~\cite{guo2021pct, zhao2021point, lim2021planar}.
In projection-based methods, 3D shapes are often projected into multiple views to obtain comprehensive features. Then they are fused with well-designed weight.
The most common instances of such a method are MVCNN~\cite{MVCNN, zhang2016health, zhang2023flowbot++} which employs a straightforward technique of max-pooling to obtain a global descriptor from multi-view features, and MHBN~\cite{yu2018multi, zhang2020dex, yao2023apla} that makes use of harmonized bilinear pooling to integrate local convolutional features and create a smaller-sized global descriptor.
Among all the point-based methods, PointNet~\cite{qi2017pointnet, zhang2023flowbot++} is a seminal work that directly takes point clouds as its input and achieves permutation invariance with a symmetric function. Since features are learned independently for each point in PointNet~\cite{qi2017pointnet}, the local structural information between points is hardly captured. Therefore, PointNet++~\cite{qi2017pointnet++, eisner2022flowbot3d} is proposed to capture fine geometric structures from the neighborhood of each point. 

Recently, several 3D large pre-trained models are proposed. CLIP~\cite{clip} is a cross-modal text-image pre-training model based on contrastive learning. PointCLIP~\cite{pointclip} and PointCLIP v2~\cite{pointclipv2} extend CLIP’s 2D pre-trained knowledge to 3D point cloud understanding. Some other 3D processing methods based on large pre-trained model CLIP are proposed in recent years, including CLIP2Point~\cite{huang2022clip2point}, CLIP$^2$~\cite{zeng2023clip}, and CG3D~\cite{CG3D}. Similar to some of those methods, our model combines widely used point cloud learning architectures with novel large pre-trained models such as CLIP. The key difference here is that our point-based network hierarchically guides the generation of prompts by using direct 3D point cloud features, which largely reduces the domain gap between training and testing data for better zero-shot and few-shot learning. We also design a simpler yet more effective projection method, an intuitive multi-view fusion strategy, to make the projection processing as efficient as possible for the zero-shot task.

\subsection{Domain Adaptation and Generalization in Vision-Language Models}
Vision-language pre-trained models such as CLIP exhibit impressive zero-shot generalization ability to the open world. Despite  the good zero-shot performance, it is found that further adapting CLIP using task-specific data comes at the expense of out-of-domain (OOD) generalization ability.~\cite{radford2021learning, wortsman2022robust, shen2024diffclip}.
Recent advances explore improving the OOD generalization of CLIP  on the downstream tasks by adapter learning 
~\cite{gao2021clip, zhang2021tip, avigal20206}, model ensembling~\cite{wortsman2022robust, elmquist2022art}, test-time adaptation~\cite{shu2022test}, prompt learning 
~\cite{zhou2022learning, CoCoOp}, and model fine-tuning~\cite{shu2023clipood}. Specifically, in StyLIP~\cite{bose2023stylip, devgon2020orienting}, style features are extracted hierarchically from the image encoder of CLIP to generate domain-specific prompts. To the best of our knowledge, our work is the first to extract style features of 3D point clouds directly from point-based processing networks. With the extracted style features, we are able to generate domain-specific prompts for the text encoder of  CLIP.

\subsection{Style Transfer using Diffusion Model}
Traditional style transfer is one of the domain generalization methods which helps reduce the domain gap between source and  target domains by aligning their styles while preserving the content information~\cite{3D-style-transfer}. For example, Domain-Specific Mappings~\cite{chang2020domain} has been proposed to transfer the style of source domains to a target domain without losing domain-specific information. 
Fundamentally different from traditional methods of style transfer that operate on pre-existing images, text-to-image diffusion models generate an image based on a textual description that may include style-related information, thus embedding the style defined in textual descriptions into the generated images. These models are based on a diffusion process, which involves gradually smoothing out an input image to create a stylized version of it, showing a strong ability in capturing complex texture and style information in images.  For instance, Glide~\cite{nichol2021glide}, Disco Diffusion~\cite{dhariwal2021diffusion}, Stable Diffusion 
~\cite{rombach2022high}, and Imagen~\cite{NEURIPS2022_ec795aea} all support text-to-image generation and editing via a diffusion process, which can further be used to change the image styles via textual prompts variations.
  We refer to the textual prompts in diffusion models as \textbf{style prompts} as they often encapsulate the style of generated images. In our work, we implement different text-to-image diffusion models and choose ControlNet~\cite{controlnet} to help transfer depth map from 3D projection to a more realistic image style, minimizing its style gap with the training images of CLIP.

\begin{figure*}[ht]
\centering
\includegraphics[width=350pt]{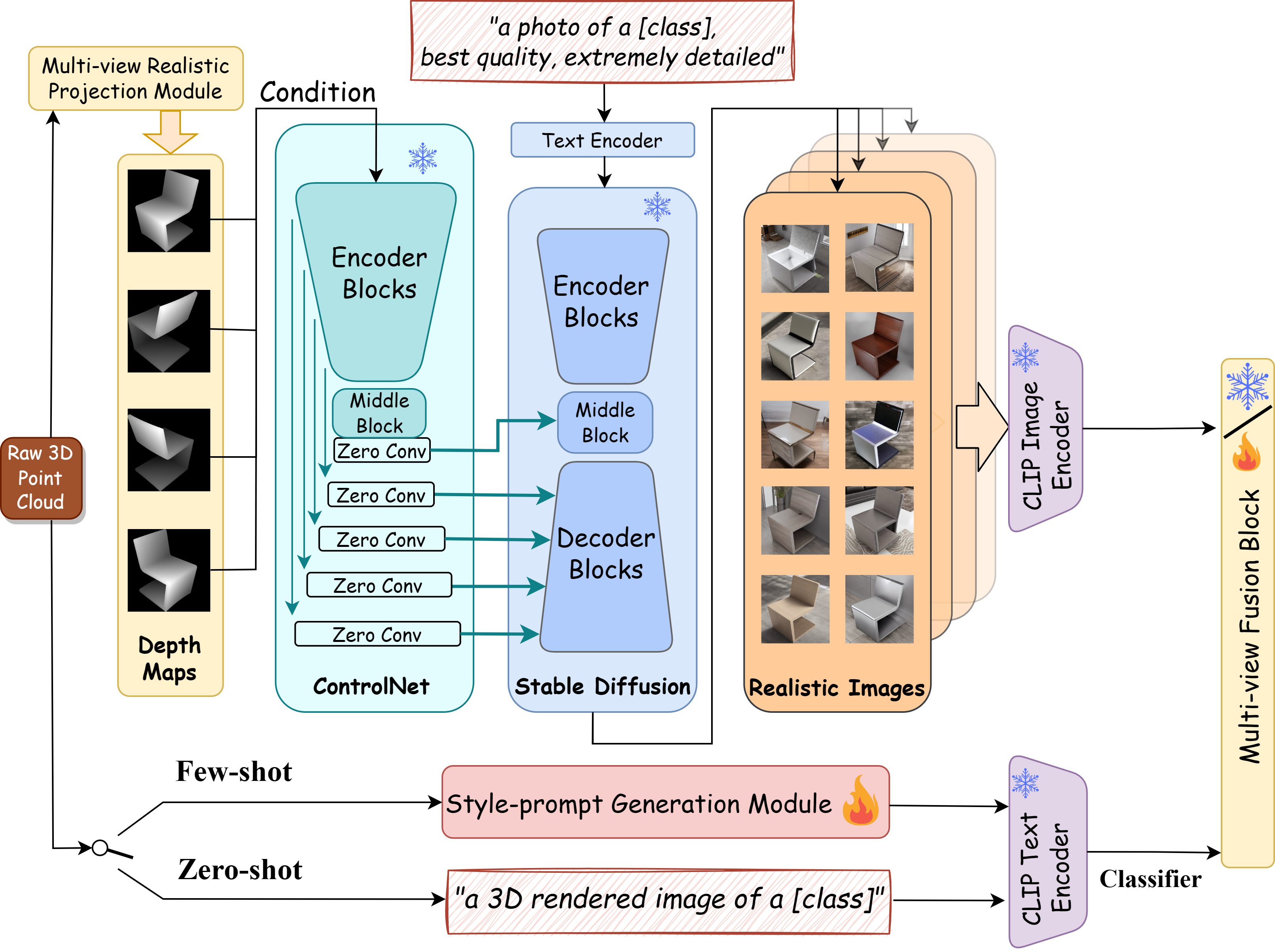}
\caption{\textbf{Framework Structure of DiffCLIP.} {In the visual branch, DiffCLIP has two modules: Multi-View Realistic Projection Module, which produces multi-view depth maps, and Stable-Diffusion-Based Style Transfer Module, which uses a pre-trained ControlNet and frozen Stable Diffusion network to transfer styles on the depth images. In the textual branch, DiffCLIP uses an optional Style-Prompt Generation Module for few-shot tasks and manual prompts for zero-shot tasks. Frozen CLIP image encoder and text encoder are used to generate feature representations of images and text which then go through a Multi-Modal Fusion Block.}}
\label{fig:framework}
\end{figure*}

\section{Method}
DiffCLIP is a novel neural network architecture that enhances the performance of the CLIP model on 3D point processing tasks via  style transfer and style-prompt generation.
In Section~\ref{revisit}, we first revisit PointCLIP~\cite{pointclip, jin2024multi} and PointCLIP v2~\cite{pointclipv2, pan2023tax} that first introduce the CLIP model into the processing of 3D point clouds. Then in Section~\ref{framework}, we introduce the framework and motivation of each part of DiffCLIP, which minimizes the domain gap between 3D tasks and 2D pre-training data. In Section~\ref{using our model}, we present the detailed usage of DiffCLIP on zero-shot and few-shot classification tasks.

\subsection{Revisiting PointCLIP and PointCLIP v2}
\label{revisit}
PointCLIP~\cite{pointclip} is a recent extension of  CLIP, which enables zero-shot classification on point cloud data.
It involves taking a point cloud and converting it into multiple depth maps from different angles without rendering. The predictions from the individual views are combined to transfer knowledge from 2D to 3D. Moreover, an inter-view adapter is used to improve the extraction of global features  and fuse the few-shot knowledge acquired from 3D into  CLIP  that was pre-trained on 2D image data. 

Specifically, for an unseen dataset of $K$ classes, PointCLIP constructs the textual prompt by placing all category names into a manual template, termed prompt, and then encodes the prompts into a $D$-dimensional textual feature, acting as the zero-shot classifier $W_t\in \mathbb{R}^{K\times D}$. Meanwhile, the features of each projected image from $M$ views are encoded as ${f_i}$ for $i=1,...,M$ by the visual encoder of CLIP. On top of this, the classification $\mathrm{logits}_i$ of view $i$ and the final $\mathrm{logits}_p$ of each point are calculated by
\begin{subequations}
\label{logits of pointclip}
\begin{minipage}{.46\textwidth}
    \begin{equation}
    \mathrm{logit} s_i=f_iW_t^T, \forall i=1,...,M
    \end{equation}
\end{minipage}%
\newline
\begin{minipage}{.46\textwidth}
    \begin{equation}
    \mathrm{logit} s_p=\sum_{i=1}^M \alpha_i \mathrm{logit}s_i,
\end{equation}
\end{minipage}%
\end{subequations}

where $\alpha_i$ is a hyper-parameter weighing the importance of view $i$.

PoinCLIP v2~\cite{pointclipv2} introduces a realistic shape projection module to generate more realistic depth maps for the visual encoder of CLIP, further narrowing down the domain gap between projected point clouds and natural images. Moreover,  a large-scale language model, GPT-3~\cite{gpt-3}, is leveraged to automatically design a more descriptive 3D-semantic prompt for the textual encoder of CLIP instead of the previous hand-crafted one.

\subsection{DiffCLIP Framework}
\label{framework}

The  DiffCLIP framework is illustrated in Fig.~\ref{fig:framework}. We describe each component of DiffCLIP in detail in the following sections.

\subsubsection{Multi-View Realistic Projection}
\label{}
In DiffCLIP, we use stable diffusion~\cite{stablediffu} to help transfer projected depth maps to a more realistic image style, minimizing the domain gap with the training images of CLIP. To generate realistic depth maps for better controlling the stable diffusion model, as well as to save computational resources  for zero-shot and few-shot tasks, we design a multi-view realistic projection module, which has three steps: proportional sampling, central projection, and 2D max-pooling densifying.

\textbf{Portional Sampling.}
We randomly sample points on each face and edge of the object in the dataset, the number of which is proportional to the area of the faces and the length of the edges. Specifically, assuming that $k_l$ points are sampled on a line with length $l$ and $k_f$ points are sampled on a face with area $s$, we set sampling threshold values $\beta_1$ and $\beta_2$, so the number of sampled points can be calculated by $k_l=\frac{l}{\beta_1}$ and $k_f=\frac{s}{\beta_2}$.

\textbf{Central Projection.}
For each projection viewpoint $n\in\{1,2,...,N\}$, we select a specific projection center and projection plane, and use affine transformation to calculate the coordinates of the projection point on the projection plane for each sampling point. The pixel intensity $d$ is calculated as $d=\frac{1}{dis}$, where $dis$ is the distance between the sampling point and our projection plane. On the 2D plane, the projected point may not fall exactly on a pixel, so we use the nearest-neighbor pixel to approximate its density.


\textbf{2D Max-pooling Densifying.}
In the central projection step, the nearest-neighbor pixel approximation may yield many pixels to be unassigned, causing the projected depth map to look unrealistic, sparse, and scattered. To tackle this problem, we densify the projected points via a local max-pooling operation to guarantee visual continuity. For each projected pixel of the depth map, we choose the max density $d_{max}$ among four points: the pixel itself and the pixels to its right, bottom, and bottom right. We then assign the max density to the neighboring four pixels.

\begin{figure}[ht]
\centering
\includegraphics[width=\linewidth]{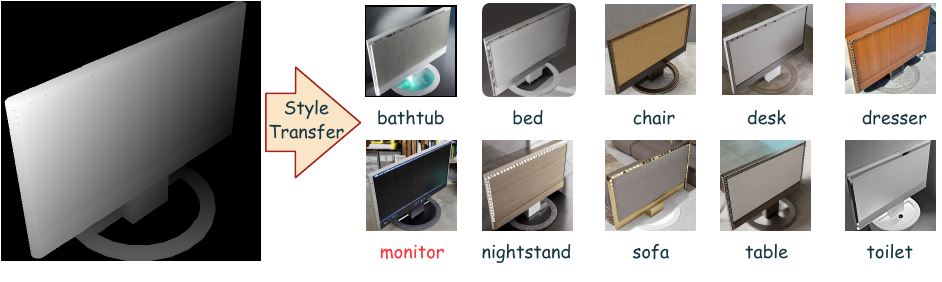}
\caption{\textbf{Style Transfer using Stable Diffusion and ControlNet:} Illustrating Results for 10 Categories (right) in the ModelNet10 Dataset Using a ``Monitor'' Depth Map (left). For example, when transferring the style of "monitor" to "bathtub", the base of the monitor will be filled with blue rippling water. When transferring the style of "monitor" to "chair" or "sofa", characteristic textures of these objects are displayed. When using the "monitor" label to transfer its own style, the resulting image clearly generates the toolbar and menu bar on the computer screen.}
\label{fig:DiffuExample}
\end{figure}

\subsubsection{Stable-Diffusion-Based Style Transfer}
To better use stable diffusion~\cite{stablediffu} in our task-specific domain, where the input is depth maps, we use ControlNet~\cite{controlnet}, a robust neural network training method to avoid overfitting and to preserve generalization ability when large models are trained for specific problems. {In DiffCLIP, ControlNet employs a U-Net architecture identical to the one used in stable diffusion. It duplicates the weights of an extensive diffusion model into two versions: one that is trainable and another that remains fixed. The linked trainable and fixed neural network units incorporate a specialized convolution layer known as ``zero convolution." In this arrangement, the weights of the convolution evolve from zero to optimized parameters in a learned manner.} ControlNet has several implementations with different image-based conditions to control large diffusion models, including Canny edge, Hough line, HED boundary, human pose, depth, etc. We use the frozen pre-trained parameters of ControlNet under depth condition, which is pre-trained on 3M depth-image-caption pairs from the internet. The pre-trained ControlNet is then used to generate our own realistic images of depth maps using different class labels as prompts. We illustrate the style-transfer effects of ControlNet in Fig.~\ref{fig:DiffuExample}.

\subsubsection{Style-Prompt Generation
}
\label{Style-Prompt Generation from Point-based Network}
\begin{figure*}
\begin{center}
\includegraphics[width=\textwidth]{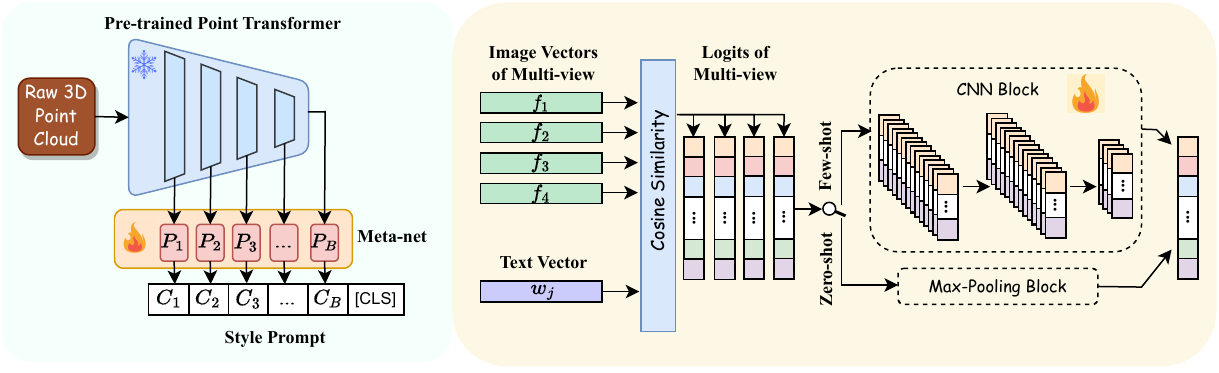}
\end{center}
\caption{\textbf{Prompt generation module} (left) and \textbf{Multi-view Fusion Block} (right) of DiffCLIP.}
\label{fig:submodule}
\end{figure*}

Optionally, when we want to do few-shot learning, we set up a style prompt generation module to embed style into prompts as is shown in the left part of Fig.~\ref{fig:submodule}. 

We make use of ``style features'' in point cloud processing analogous to the usage in 2D vision. In 2D vision tasks, ConvNets act as visual encoders that extract features at different layers: higher-level layers tend to learn more abstract and global features while lower-level layers detect simple and local features such as edges, corners, and blobs~\cite{zeiler2014visualizing, ResNet}. We conjecture that in the 3D point cloud case, the combination of features extracted from multiple levels could also improve the generalizability of the model, similar to how style features improve learning on RGB images in 2D vision tasks.

DiffCLIP leverages CLIP's frozen vision encoder $f_v$ and text encoder $f_t$. Additionally,  a pre-trained point-based network, Point Transformer $f_p$~\cite{zhao2021point}, is used in DiffCLIP. The Point Transformer operates by first encoding each point in the point cloud into a high-dimensional feature vector using a multi-layer perception network~\cite{werbos1974beyond}, and then processing the encoded  feature vectors are then processed by a series of self-attention layers, similar to the ones used in the Transformer~\cite{transformer} architecture. 

We pre-train the Point Transformer network using our customized dataset, ShapeNet37, which consists of sampled data from ShapeNetCore~\cite{chang2015shapenet}. The network comprises several blocks, each generating feature maps at different scales. To incorporate domain-specific characteristics, we project the output  features from each block into a token and then concatenate them together with a class label to generate the style prompt as the input of CLIP's text encoder.

\subsection{Using DiffCLIP}
\label{using our model}
\subsubsection{Zero-shot 3D classification}
\label{using our model_zero-shot}
For each 3D point cloud input $x$, after being projected into $M$ views and densified, we obtain their smooth depth maps $x^{\prime}_i,\;
\forall i\in\{1,...,M\}$. These depth maps are then fed into ControlNet as conditions to guide the image generation of stable diffusion. We use the prompt ``\texttt{a photo of a [class], best quality, extremely detailed}'' as the default style prompt input to the stable diffusion model, denoted as $t_j$, where $j\in\{1,...,K\}$. The generated realistic images are represented as $x^{\prime\prime}_{ij}=d(x^{\prime}_i, t_j)$, where the function $d(\cdot)$ denotes the stable diffusion combined with  ControlNet. Based on the realistic images generated by the projection and style transfer stages, we employ CLIP to extract their visual features $f_{ij}=f_v(x^{\prime\prime}_{ij})$. In the textual branch, we use ``\texttt{a photo of a [class]}'' as the prompt and encode their textual features as the zero-shot classifier $W_t\in \mathbb{R}^{K\times D}$. Furthermore, the classification $\mathrm{logit}s_{ij}$ for each view $i$ and each style transfer text guidance $j$ are calculated separately as follows:

\begin{equation}
\mathrm{logit}s_{ij}=f_{ij}W_t^T, \forall i\in\{1,...,M\},\;j\in\{1,...,K\}
\end{equation}

For each style transfer text guidance $j$, the classification $\mathrm{logit}s_{j} \in \mathbb{R}^{1\times K}$ is defined as
$MaxP([\mathrm{logit}s_{1j}; ... ;\mathrm{logit}s_{Mj}])$, where $MaxP(\cdot)$ represents taking the maximum value of each column in the matrix. Then, the probability matrix $P \in \mathbb{R}^{K \times K }$  is generated by $P=[softmax(\mathrm{logit}s_{1}); ... ;softmax(\mathrm{logit}s_{K})]$. 

We design two strategies to calculate the final probability vector $\mathbf{p}= [p_1,p_2,...,p_K] \in \mathbb{R}^{K}$ for each classification. Each of them includes a global logits part and a local one in order to make full use of raw probability distributions of each diffusion result from each view on all the classes. The first strategy is given by 

\begin{equation}
    \mathbf{p} =\beta_1 \mathbf{p_{glo}}+\beta_2 \mathbf{p_{loc}}
\label{equ:3}
\end{equation}
where 
\begin{align}
    \mathbf{p_{glo}}&=\left[\sum_{i=1}^K P_{i1}\mathbb{I}(P_{i1}\leq P_{11}),...,\sum_{i=1}^K P_{iK}\mathbb{I}(P_{iK}\leq P_{KK})\right]\\
    \mathbf{p_{loc}}&=\left[P_{11},...,P_{KK}\right]
\label{equ:5}
\end{align}

and $\beta_1$ and $\beta_2$ are hyper-parameters. $\mathbf{p_{glo}}$ represents summing all values in the matrix that are no more than the diagonal by column, which represents the global information, and $\mathbf{p_{loc}}$ returns the diagonal entries of $P$ that represent the probabilities of the realistic images generated by text guidance $j$ being classified into category $j$, which provides local information. We illustrate the detailed computation in our supplementary material.
The second strategy to aggregate the matrix $P$ to calculate the final logits is as follows:

\begin{equation}
\mathbf{p} = \mathrm{norm}(\mathbf{p_{glo}}) * \mathbf{p_{loc}}
\label{equ:6}
\end{equation}
where
\begin{align}
\begin{split}
\mathbf{p_{glo}} &= \left[\sqrt[K]{\Pi_{i=1}^{K}{P_{i1}}}, ..., \sqrt[K]{\Pi_{i=1}^{K}{P_{iK}}}\right] \\
\mathbf{p_{loc}} &= \left[\max_{i \in \{1,2,...,K\}}{P_{i1}}, ..., \max_{i \in \{1,2,...,K\}}{P_{iK}}\right] \\
\mathrm{norm}(\boldsymbol{p}) &= (\boldsymbol{p}-\min(\boldsymbol{p}))/(\max(\boldsymbol{p})-\min(\boldsymbol{p}))
\end{split}
\end{align}
Each entry in $\mathbf{p_{glo}}$ represents the  global feature of each category, which is defined as the geometric mean of the probabilities of that category across all style transfer results. Each entry in $\mathbf{p_{loc}}$ represents the local feature for each category and is obtained from the diffusion result that is most similar to that category itself. Detailed clarification of these equations are shown in the appendix. The experiment results demonstrate that these two calculation strategies have their own strengths and weaknesses on different datasets.

\begin{figure}[]
\centering
\includegraphics[width=210pt]{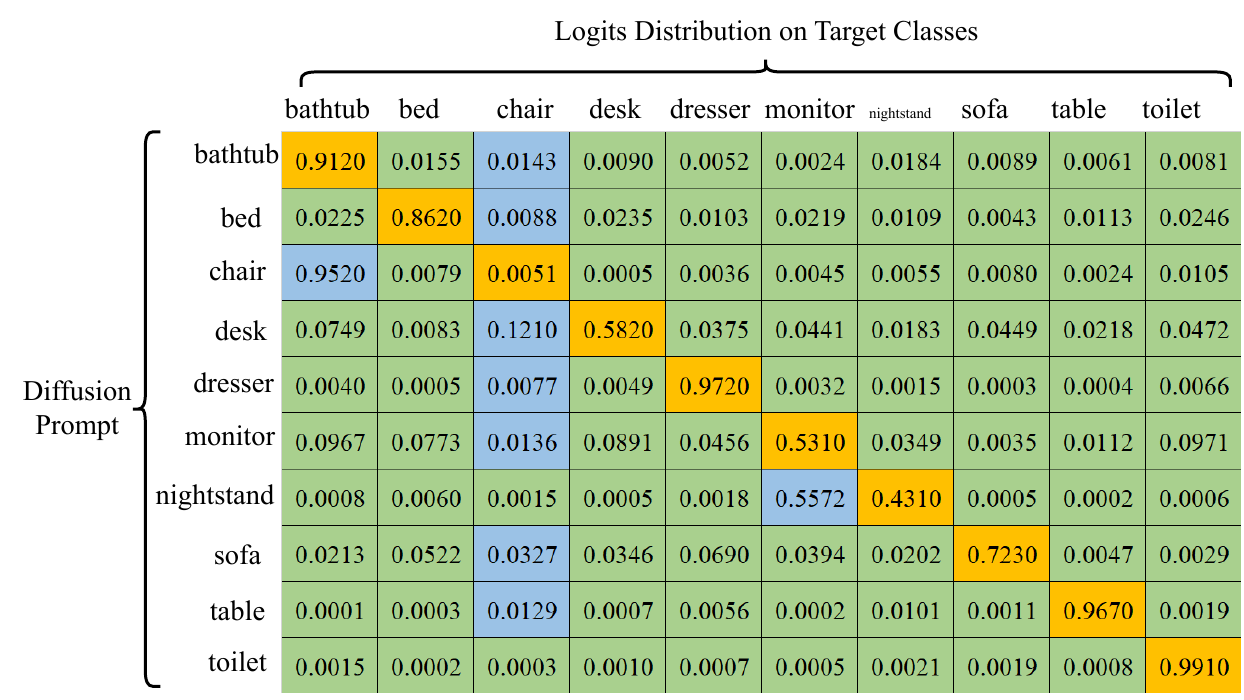}
\caption{An example of matrix $P$.}
\label{fig:appd_calc}
\end{figure}

To better illustrate the calculation process of probability matrix $P$ mentioned in equation~\ref{equ:3} and equation~\ref{equ:5}, we give the matrix in Fig.~\ref{fig:appd_calc} as an example. In equation~\ref{equ:6}, while calculating $\mathbf{p_{glo}}$, numbers in blue boxes, which are bigger than numbers in orange boxes, are ignored. 

\subsubsection{Few-Shot 3D Classification}
In the context of few-shot classification tasks, a primary innovation of our model lies in the generation of style prompts. The image branch of the model remains identical to the one used in zero-shot classification tasks. As detailed in Section~\ref{Style-Prompt Generation from Point-based Network}, we assume that the point-based network consists of $B$ blocks. To incorporate domain-specific characteristics, we establish $B$ meta-nets as style projectors ${\mathcal{P}_b(\cdot;\theta_b)}\;, \forall b\in\{1,...,B\}$ to encode domain characteristics into $B$ prefix features ${c_b}$, where $\theta_b$ represents the parameters of the $b$-th meta-net. Assuming that the textual feature from the $b$-th block is $D_b$-dimension, the total dimension of our textual encoder is $D=\sum_{b=1}^B D_b$. Specifically, $c_b \in \mathbb{R}^{D_b}$ is computed by $c_b(x)=\mathcal{P}_b(\mathcal{F}_b(x);\theta_b)$.

To generate a style-prompt for a given $(x, y)$ pair, where $x\in \mathbb{R}^{2048\times3}$ is an original 3D point cloud input, we define the full style prompt $t_y$ by

\begin{equation}
t_{y}=[c_1(x)][c_2(x)]...[c_B(x)][CLS_y],
\end{equation}

where $[CLS_y]$ is the word embedding of label $y$. Subsequently, a zero-shot classifier $W_{t}\in \mathbb{R}^{D\times K}$ can be generated.

Following the computation of logits, as in zero-shot classification, we establish another trainable module, using a convolutional neural network to fuse logits $s_i$ of each view $i\in\{1,...,M\}$, illustrated in the right portion of Fig.~\ref{fig:submodule}.

\section{Experiments and Results}

\begin{table}
\centering
\begin{threeparttable}
\renewcommand\arraystretch{1.3}
\resizebox{\linewidth}{!}{
\begin{tabular}{l|ccccc}
\toprule
\multicolumn{1}{c|}{}                         & \multicolumn{5}{c}{Zero-shot Performance}                                                       \\ \cline{2-6} 
\multicolumn{1}{c|}{}                         &                              &                              & \multicolumn{3}{c}{ScanObjectNN}  \\ \cline{4-6} 
\multicolumn{1}{c|}{\multirow{-3}{*}{Method}} & \multirow{-2}{*}{MN10} & \multirow{-2}{*}{MN40} & OBJ\_ONLY & OBJ\_BG & PB\_T50\_RS \\ \hline
PointCLIP~\cite{pointclip}                     & 30.2                         & 23.8                         & 21.3      & 19.3    & 15.4        \\
Cheraghian~\cite{pointclipv2}                  & 68.5                         & -                            & -         & -       & -           \\
CLIP2Point~\cite{huang2022clip2point}          & 66.6                         & 49.4                         & 35.5      & 30.5    & 23.3        \\
PointMLP+CG3D~\cite{CG3D}                      & 64.1                         & 50.4                         & - & - & 25.0          \\
PointTransformer+CG3D~\cite{CG3D}              & 67.3                         & 50.6                         & - & - & 25.6          \\
PointCLIP v2~\cite{pointclipv2}                & 73.1                         &\textbf{64.2}                 & \textbf{50.1}& 41.2    & \textbf{35.4}
     \\
ReCon~\cite{ReCon}                           & 75.6                          &61.7                          & 43.7      &40.4    & 30.5
     \\ \hline
\textbf{DiffCLIP}                            & \textbf{82.4}                 & 54.2                         & 45.3           &\textbf{43.2}   & 35.2            \\
\bottomrule
\end{tabular}
}
\captionsetup{width= .46\textwidth}
\captionsetup{hypcap=false}
\caption{Zero-shot classification accuracy (\%) of DiffCLIP on ModelNet10~\cite{ModelNet}, ModelNet40~\cite{ModelNet}, ScanObjectNN~\cite{ScanobjectNN}.}
\label{tab:zero-shot}
\end{threeparttable}
\end{table}

\subsection{Zero-Shot Classification}

We evaluate the zero-shot classification performance of DiffCLIP on three well-known datasets: ModelNet10~\cite{ModelNet, avigal20206, avigal2021avplug}, ModelNet40~\cite{ModelNet}, and ScanObjectNN~\cite{ScanobjectNN}. For each dataset, we require no training data and adopt the full test set for evaluation. For the pre-trained CLIP model, we adopt ResNet-50~\cite{ResNet} as the visual encoder and transformer~\cite{transformer} as the textual encoder by default, and then try several other encoders in the ablation studies. We initially set the prompt of stable diffusion as ``\texttt{a photo of a [class], best quality, extremely detailed}'', and then we try multiple prompts to adapt the three different datasets later. 

\begin{table*}
\begin{center}
\small
\renewcommand\arraystretch{1.2}
\resizebox{\textwidth}{!}
{
\begin{tabular}{lcccccccccccc}
\toprule
                                & bathtub & bed  & chair & desk & dresser & monitor & nightstand & sofa & table & toilet & Average \\ \hline
DiffCLIP                        & 97.0    & 88.4 & 46.5  & 33.7 & 94.0    & 99.0    & 99.0       & 78.0 & 87.0  & 94.0   & 82.2    \\
Ablation & 83.0    & 62.8 & 3.5   & 29.1 & 36.0    & 100.0   & 98.0       & 98.0 & 89.0  & 80.0   & 71.4   \\
\bottomrule
\end{tabular}
}
\end{center}
\caption{Ablation study. Line 2 shows zero-shot classification accuracy(\%) of DiffCLIP without ControlNet and Stable diffusion module on ModelNet10.}
\label{tab:ablation}
\end{table*}

\textbf{Performance.}
In Table~\ref{tab:zero-shot}, we present the performances of zero-shot DiffCLIP for three datasets with their best performance settings. Without any 3D training data, DiffCLIP achieves an accuracy of 43.2\% on OBJ\_BG of ScanObjectNN dataset, an accuracy of 82.4\% for zero-shot classification on ModelNet10, which are state-of-the-art performance, and an accuracy of 35.2\% for zero-shot classification on PB\_T50\_RS, which is comparable to state-of-the-art performace. While PointCLIP v2's~\cite{pointclipv2} results are sometimes better than ours, GPT-3 is used in the PointCLIP v2 model, which would require careful manual labor and further selection of the prompts that are useful for each sample in the test set. DiffCLIP does not require extensive manual tweaking of the prompts to achieve acceptable results by leveraging stable diffusion, which realizes full zero-shot classification without overfitting to each test set.

\textbf{Ablation Studies.}
We conduct ablation studies on zero-shot DiffCLIP concerning the importance of stable diffusion plus ControlNet module, as well as the influence of different projection views and projection view numbers on ModelNet10. For zero-shot DiffCLIP structure without style transfer stages, in which the depth maps are directly sent into the frozen image encoder of CLIP, the best classification accuracy reaches 58.8\%, while implementing ViT/16 as the image encoder of CLIP. Hence, we can see the style transfer stage increases the accuracy by 21.8\%. Detailed ablation results are shown in Table~\ref{tab:ablation}. 

In terms of the number of projected views, we try 1 (randomly selected), 1 (elaborately selected), 4, and 8 projection view(s) on ModelNet10. In Table~\ref{tab:mult-view}, records are taken from a bird's-eye view angle of 35 degrees, and the angles in the table represent the number of degrees of horizontal rotation around the Z-axis. The result of one elaborately selected view from a -135° angle outperforms others. Due to the characteristics of ModelNet10 dataset that all the objects are placed horizontally, projection from 4 views does not show obvious advantages. But the performance of the method on ModelNet40 is improved from multi-view projection because objects have different spatial arrangements, as shown in Table~\ref{tab:40-multiview}. We also ablate on the choice of different visual backbones, shown in Table~\ref{tab:visual encoders}. The results suggest that ViT/16 yields the best results in our framework.

\textbf{Prompt Design.} In DiffCLIP, we design several different prompts as both the textual branch of the CLIP model and the style prompt input of stable diffusion. For ModelNet10 and ModelNet40, we implement a default prompt for stable diffusion. The performance of multiple prompt designs for CLIP on ModelNet10 is shown in Table~\ref{tab:prompt design}.
We try to place the class tag at the beginning, middle, and end of sentences, among which class tag located at the middle of the sentence performs best. {Specifically, real-world objects in ScanObjectNN are incomplete and unclear, we elaborately design its diffusion style prompt as ``\texttt{a photo of a [class], behind the building, best quality, extremely detailed}'', in order to generate an obstacle in front of the target object, which is shown to be more recognizable for CLIP's image encoder.}

\noindent \begin{minipage}[t]{.46\textwidth}
    \centering
\renewcommand\arraystretch{1.2}
\resizebox{\textwidth}{!}{
\begin{tabular}{@{}ccccccc@{}}
\toprule
\multicolumn{7}{c}{Different Projection View Performance}                                                                \\ \midrule
\multicolumn{1}{c|}{View}       & 1(-135°)     & 1(-45°)     & 1(45°)     & 1(135°)     & 1(random)   & 4 \\ \midrule
\multicolumn{1}{c|}{ModelNet10} & \textbf{82.4}& 67.5       & 55.2      &   77.9          & 54.3       & 80.1 \\ \bottomrule
\end{tabular}
}
\captionsetup{width=\textwidth}
\captionsetup{hypcap=false}
\captionof{table}{Zero-shot classification accuracy (\%) of DiffCLIP on ModelNet10 under various projection view(s) and view numbers, using ViT/16 as the visual encoder of CLIP.}
\label{tab:mult-view}
\end{minipage}%
\hfill
\begin{minipage}[t]{.46\textwidth}
\renewcommand\arraystretch{1.3}
\resizebox{\textwidth}{!}{
\begin{tabular}{lccccc}
\toprule
\multicolumn{6}{c}{Different Visual Encoders} \\ \hline
Models     & RN50  & RN101  & ViT/32  & ViT/16    & RN.$\times$16\\ \hline
PointCLIP  & 20.2  & 17.0   & 16.9    & 21.3            & 23.8        \\
\textbf{DiffCLIP}   & 68.5  & 73.1   & 77.3    & \textbf{82.4}   & 74.4        \\ \bottomrule
\end{tabular}
}
\captionsetup{width=\textwidth}
\captionsetup{hypcap=false}
\captionof{table}{Zero-shot classification accuracy (\%) of DiffCLIP on ModelNet10 implementing different encoders with best projection view(s).}
\label{tab:visual encoders}
\end{minipage}
\begin{table}[h]
\centering
\small
\renewcommand\arraystretch{1.2}
\begin{tabular}{lcc}
\toprule
           & Single-view & Multi-view \\ \hline
ModelNet40 & 49.7        & 54.2    \\  
\bottomrule
\end{tabular}
\caption{Zero-shot classification accuracy(\%) of DiffCLIP on ModelNet40 under different view numbers.}
\label{tab:40-multiview}
\end{table}
\noindent 
\begin{minipage}[t]{.46\textwidth}
    \centering
\renewcommand\arraystretch{1.2}
\resizebox{\linewidth}{!}{
\centering
\renewcommand\arraystretch{1.2}
\begin{tabular}{lllll}
\toprule
Prompt for CLIP                      & \multicolumn{4}{l}{Accuracy} \\ \hline
"a photo of a + [C]"           & \multicolumn{4}{l}{80.1}       \\
"a rendered image of + [C]"    & \multicolumn{4}{l}{78.1}       \\
"a 3D rendered image of + [C]" & \multicolumn{4}{l}{79.4}       \\
"[C] + with white context"     & \multicolumn{4}{l}{78.0} \\ 
"[C] + with white background"     & \multicolumn{4}{l}{79.9} \\ 
"a 3D image of a + [C] + with rendered background" & \multicolumn{4}{l}{\textbf{82.4}}\\ \bottomrule
\end{tabular}
}
\captionsetup{width=\textwidth}
\captionsetup{hypcap=false}
\captionof{table}{Zero-shot classification accuracy (\%) of DiffCLIP on ModelNet10 using different prompt for CLIP.}
\label{tab:prompt design}
\end{minipage}%
\hfill
\begin{minipage}[t]{.46\textwidth}
\resizebox{\textwidth}{!}{
\centering
\renewcommand\arraystretch{1.2}
\begin{tabular}{lcccc}
\toprule
\multirow{2}{*}{Method} & \multicolumn{4}{c}{ModelNet10}     \\ \cline{2-5}
                                    & 1-shot & 2-shot & 4-shot & 8-shot  \\ \hline
PointNet~\cite{qi2017pointnet}      & 42.9   & 44.3   & 61.4   & 78.0    \\
PointNet++~\cite{qi2017pointnet++}  & 53.2   & 69.3   & 73.9   & 83.8    \\ \hline
\textbf{DiffCLIP}                   & 45.8   & 48.2   & 62.6   & 79.0     
\\ \bottomrule
\end{tabular}
}
\captionsetup{width=\textwidth}
\captionsetup{hypcap=false}
\captionof{table}{Few-shot classification accuracy (\%) of DiffCLIP on ModelNet10 with different shot numbers.}
\label{tab:few-shot}
\end{minipage}

\subsection{Few-shot Experiments}
We experiment DiffCLIP with the trainable Style-Prompt Generation Module and the trainable Multi-view Fusion Module under 1, 2, 4, and 8 shots on ModelNet10. For $K$-shot settings, we randomly sample $K$ point clouds from each category of the training set. 
We set the learning rate as 1e-4 initially to train the meta-nets in the style prompt generation module, and then turn the learning rate of the meta-net into 1e-7 and train the multi-view fusion block. Due to the limitation of memory of our GPU, the training batch size is set as 1. Default prompts are used for stable diffusion.

\textbf{Performance.} As is shown in Table~\ref{tab:few-shot}, we compare few-shot classification performance of DiffCLIP to PointNet~\cite{qi2017pointnet} and PointNet++~\cite{qi2017pointnet++} in terms of $K$-shot classification accuracy. 
DiffCLIP surpasses PointNet in performance and approximately approaches PointNet++. It reaches an accuracy of 79.0\% under 8-shot training.

\section{Discussion and Limitations}
In zero-shot 3D classification tasks, DiffCLIP achieves state-of-the-art performance on OBJ\_BG of ScanObjectNN dataset and accuracy on ModelNet10 which is comparable to state-of-the-art. In few-shot 3D classification tasks, DiffCLIP surpasses PointNet in performance and approximately approaches PointNet++. However, our model does have some notable room for improvement. First, DiffCLIP's performance on zero-shot tasks needs more exploration, such as projecting 3D point cloud into more views, designing better functions to calculate the logits vector $\mathbf{p}$. Second, DiffCLIP's performance on few-shot tasks are still not that good which needs further improvement, such as fine-tuning on initial CLIP encoders as well as ControlNet. Moreover, the ability of DiffCLIP processing some other 3D tasks including segmentation and object detection is expected to be explored. We plan to leave these for future work.

\textbf{Limitations.} 
First, the implementation of stable diffusion introduces a time-intensive aspect to both the training and testing of the model due to the intricate computations required in the diffusion process, which can elongate the time for model execution~\ref{tab:time consumption}. We also acknowledge that the scale of our pre-training dataset for the point transformer is presently limited. This constraint impacts the performance of DiffCLIP. A larger, more diverse dataset would inherently provide a richer source of learning for the model, thereby enhancing its capability to understand more data.

\begin{table}[]
\centering
\renewcommand\arraystretch{1.2}
\begin{tabular}{ccc}
\toprule
Step for diffusion & Time consumption & Accuracy \\ \hline
0                  & 94min            & 71.7     \\
5                  & 304min           & 80.7     \\
10                 & 549min           & 82.4     \\
15                 & 862min           & 82.2     \\
20                 & 1061min          & 80.6     \\
\bottomrule
\end{tabular}
\caption{Time consumption of zero-shot inference and accuracy(\%) of DiffCLIP with various diffusion steps on ModelNet10.}
\label{tab:time consumption}
\end{table}

\vspace{-3mm}
\section{Conclusion}

In conclusion, we propose a new pre-training framework called DiffCLIP that addresses the domain gap in 3D point cloud processing tasks by incorporating stable diffusion with ControlNet in the visual branch and introducing a style-prompt generation module for few-shot tasks in the textual branch. The experiments conducted on ModelNet10, ModelNet40, and ScanObjectNN datasets demonstrate that DiffCLIP has strong abilities for zero-shot 3D understanding. The results show that DiffCLIP achieves state-of-the-art performance with an accuracy of 43.2\% for zero-shot classification on OBJ\_BG of ScanObjectNN dataset and a comparable accuracy with state-of-the-art of 82.4\% for zero-shot classification on ModelNet10. These findings suggest that the proposed framework can effectively minimize the domain gap and improve the performance of large pre-trained models on 3D point cloud processing tasks.

\clearpage
{\small
\bibliographystyle{ieee_fullname}
\bibliography{egbib}
}

\clearpage

\clearpage
\appendix

\section{Visual Encoder Details}
In our experiment, we use ViT/32, ViT/16, and RN.$\times$16 as three of the image encoders. ViT/32 is short for ViT-B/32, which represents vision transformer with 32 × 32 patch embeddings. ViT/16 has the same meaning. RN.×16 denotes ResNet-50 which requires 16 times more computations than the standard model.

\section{More Information on the DiffCLIP Pipeline}
\textbf{Densify on ScanObjectNN dataset.}
In ModelNet datasets, only coordinates of a limited set of key points and normal vectors of faces are provided. To increase the density of our data representation, we perform densification on 2D depth maps after point sampling and projection. In contrast, the ScanobjectNN dataset provides coordinates for all points, so we use a different densification method. Specifically, we calculate the $k$-nearest neighbors ($k=4$) for each point and construct triangular planes by connecting the point to all possible pairs of its neighboring points.

\section{Detailed Clarification of Equations 4, 5, 6, 7}
For equation(4), $p_{loc}$ returns the diagonal entries of the matrix that represents the probabilities of the realistic images generated by text guidance j being classified into category j. $p_{glo}$ is calculated by summing all values in the matrix that are no more than the diagonal by column, the reason for this is that the values in $j_{th}$ column could be considered as the characteristic score of all generated with category j and the value in $j_{th}$ row supposed to be the largest, so we ignore the larger values.

For equation (5), '*' represents the Hadamard product. we use the product of each element in $p_{loc}$ and $p_{glo}$ as the final probability. The function $norm (\cdot )$ could scale the values between 0 and 1.

For equation (6), $p_{loc}$ regarded the maximum value as the local feature by each column which represent the diffusion result that is most similar to that category itself. The $j_{th}$ element in $p_{glo}$ is calculated by $exp{\frac{1}{K} \sum_{i=1} ^K logP_{ij}} $, we considered all values in each column for the same reason as above. The purpose of fetching the $log$ is to convert the operation of multiplication to addition, then take $exp$ to rescale the values.

For equation (7), we directly concatenate the prefix features $c_b$ and $[CLS_y]$ instead of the prompt embeddings designed by us. $c_b(x)$ is computed by the parameters obtained from Pre-trained Point Transformer with an original 3D point cloud input $x$, the process is shown in Figure3 (left).

\section{Experimental Details}
\textbf{Construction of Pretraining Dataset} In Section 3.2.3, we mentioned that we use our customized dataset, ShapeNet37, which consists of sampled data from ShapeNetCore, to pretrain point transformer. Specifically, in order to thoroughly test the generalization ability of the DiffCLIP model, we removed all data from categories that overlapped with those in ModelNet10 and ModelNet40 datasets from the ShapeNet dataset. The remaining data belonged to 37 categories, which we named ShapeNet37.

\section{Calculating the Logits of Style Transfer}
\begin{figure*}[htbp]
\centering
\includegraphics[width=360pt]{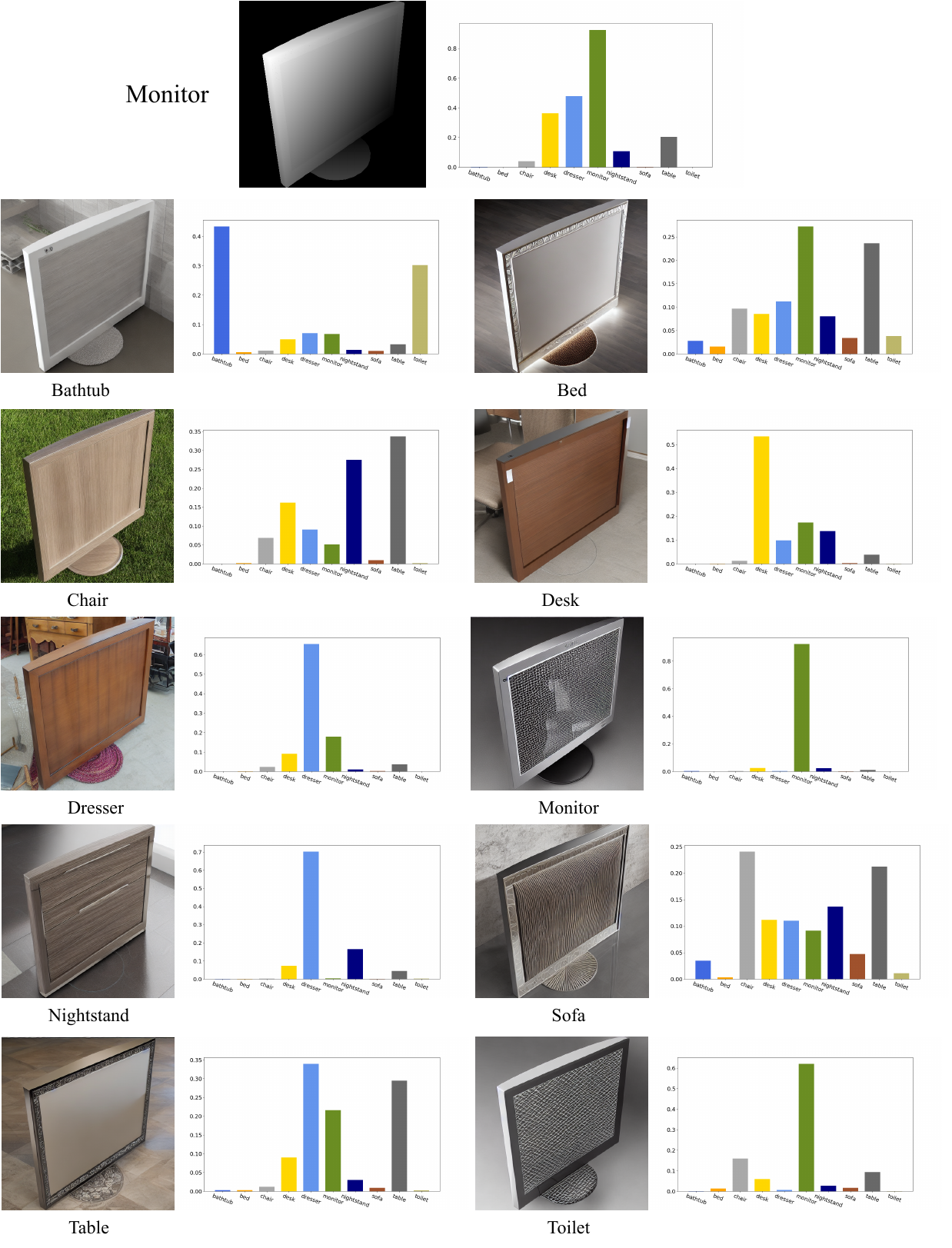}
\caption{An example of style transfer result. Logits of ten images through stable diffusion's style transfer and the following calculation from source depth condition, the 'Monitor', are shown.}
\label{fig:appd_logits}
\end{figure*}

To better illustrate the result of style transfer through stable diffusion and logits' calculation, we draw the bar chart (Fig.~\ref{fig:appd_logits}) of detailed logits of an example.

\end{document}